\documentclass{article}

\usepackage{PRIMEarxiv}

\usepackage[utf8]{inputenc} 
\usepackage[T1]{fontenc}    
\usepackage{hyperref}       
\usepackage{url}            
\usepackage{booktabs}       
\usepackage{amsfonts}       
\usepackage{nicefrac}       
\usepackage{microtype}      
\usepackage{lipsum}
\usepackage{fancyhdr}       
\usepackage{graphicx}       
\usepackage[title]{appendix}
\usepackage[square,numbers]{natbib}
\usepackage{float}
\usepackage{amsthm}
\usepackage{caption}
\usepackage[framemethod=tikz]{mdframed}
\usepackage{etoolbox} 
\usepackage{orcidlink}
\graphicspath{{media/}} 

\theoremstyle{definition}
\newmdtheoremenv[
  hidealllines=true,
  leftline=true,
  innerleftmargin=10pt,
  innerrightmargin=10pt,
  skipabove=10pt,
  skipbelow=10pt,
]{story}{Story}

\newmdtheoremenv[
  hidealllines=true,
  leftline=true,
  innerleftmargin=10pt,
  innerrightmargin=10pt,
  skipabove=10pt,
  skipbelow=10pt,
]{prompt}{Prompt}

\newmdtheoremenv[
  hidealllines=true,
  leftline=true,
  innerleftmargin=10pt,
  innerrightmargin=10pt,
  skipabove=10pt,
  skipbelow=10pt,
]{output1}{Output}

\newmdtheoremenv[
  hidealllines=true,
  leftline=true,
  innerleftmargin=10pt,
  innerrightmargin=10pt,
  skipabove=10pt,
  skipbelow=10pt,
]{pattern}{Pattern}

\newtheorem*{premise}{}
\surroundwithmdframed[
  hidealllines=true,
  leftline=true,
  innerleftmargin=10pt,
  innerrightmargin=10pt,
  skipabove=10pt,
  skipbelow=10pt,
]{premise}

\makeatletter
\renewenvironment{premise}[1][]{\par
  \normalfont \topsep6\p@\@plus6\p@\relax
  \trivlist
  \item[
        \itshape
    #1]\ignorespaces
}{%
  \popQED\endtrivlist\@endpefalse
}
\makeatother
\title{Multigenre AI-powered Story Composition}

\author{
  Edirlei Soares de Lima \orcidlink{0000-0002-2617-3394}\\
  Academy for AI, Games and Media \\
  Breda University of Applied Sciences \\
  Breda, The Netherlands\\
  \texttt{soaresdelima.e@buas.nl} \\
  \And
  Margot M. E. Neggers \orcidlink{0000-0003-1144-7580} \\
  Academy for AI, Games and Media \\
  Breda University of Applied Sciences \\
  Breda, The Netherlands\\
  \texttt{neggers.m@buas.nl} \\
  \And
  Antonio L. Furtado \orcidlink{0000-0003-3710-624X}\\
  Department of Informatics \\
  PUC-Rio \\
  Rio de Janeiro, Brazil\\
  \texttt{furtado@inf.puc-rio.br} \\
}

\begin{document}
\maketitle

\begin{abstract}
This paper shows how to construct \textit{genre patterns}, whose purpose is to guide interactive story composition in a way that enforces \textit{thematic consistency}. To start the discussion we argue, based on previous seminal works, for the existence of five \textit{fundamental genres}, namely \textit{comedy}, \textit{romance} – in the sense of epic plots, flourishing since the twelfth century –, \textit{tragedy}, \textit{satire}, and \textit{mystery}. To construct the patterns, a simple two-phase process is employed: first retrieving examples that match our genre characterizations, and then applying a form of \textit{most specific generalization} to the groups of examples in order to find their commonalities. In both phases, AI agents are instrumental, with our PatternTeller prototype being called to operate the story composition process, offering the opportunity to generate stories from a given premise of the user, to be developed under the guidance of the chosen pattern and trying to accommodate the user’s suggestions along the composition stages.
\end{abstract}

\keywords{Genres \and Types \and Variants \and Thematic Consistency \and Story Composition \and Narrative Patterns \and Most Specific Generalization \and Intelligent Agents}

\section{Introduction}

This paper proposes a strategy for preserving thematic consistency in interactive narrative composition, by letting the composition process be guided by narrative structures associated with the chosen fundamental \textit{genre}. The present work is a continuation of our research project \cite{DELIMA201631,10.1007/978-3-319-24589-8_22,LimaTypes2015} on the creation and usage of \textit{narrative patterns}, originally applied to the specification of types, in the sense of the \textit{Aarne-Thompson-Uther Index} \cite{Uther2011}. 

Countless lists of narrative genres have been publicized, with widely different numbers of items,\footnote{\url{http://www.flls.org/wp-content/uploads/2013/10/NoveList-Genre-Guide-2017.pdf}} one well-known example being the IMDB list of no less than 28 movie genres.\footnote{\url{https://help.imdb.com/article/contribution/titles/genres/GZDRMS6R742JRGAG}} The question remained whether there could be a smaller more manageable number of practically \textit{fundamental} genres, able to cover a vast majority of interactive story composition ventures and, equally important, of narrative-based video games. 

A highly reputed study is Marie-Laure Ryan’s work \cite{10.1007/978-3-540-89454-4_2}, in which she distinguishes just three types of plots and discusses how they can be appropriately handled in interactive stories and games: 1. \textit{epic}, 2. \textit{dramatic} (comprising \textit{tragedy} and \textit{comedy}), and 3. \textit{epistemic} (involving \textit{mystery} narratives). The first two date back to Aristotle, whereas the third only began in the nineteenth century. While epic plots, famously epitomized in Campbell’s \textit{Monomyth} \cite{Campbell_2008}, mostly contain physical actions, and dramatic plots deal with networks of human relations, epistemic plots are driven by the desire to know and actually consist of two narratives: events occurring in the past and events of the investigation to \textit{discover} those past events – typically having a detective or an occult sciences expert as the protagonist. 

Another famous study, equally relevant to our search for fundamental genres, was the \textit{theory of myths} exposed by the late literary scholar Northrop Frye \cite{Frye}, who ended up distinguishing four genres: 1. \textit{comedy}, 2. \textit{romance}, 3. \textit{tragedy}, and 4. \textit{satire}.

For our own study, we decided to combine Ryan’s and Frye’s proposals, by considering five fundamental genres: 1. \textit{comedy}, 2. \textit{romance} (Frey’s term, corresponding to Ryan’s epic plots), 3. \textit{tragedy}, 4. \textit{satire}, and 5. \textit{mystery}. The fifth genre has a growing prestige, constantly featuring new detectives and saviors, while, at first, we were in doubt about satire (associated with \textit{irony}), which evokes unpalatable \textit{dystopian} worlds – but to our surprise we found that \textit{dystopian games} are becoming quite popular.\footnote{\url{https://en.wikipedia.org/wiki/Category:Dystopian_video_games}} Having made this choice, we proceeded to the structural construction of comprehensive \textit{genre patterns}.

The continuation of the paper is organized as follows. Section \ref{sec2} is a brief survey of related works. The creation of patterns in correspondence with the five fundamental genres is the object of Section \ref{sec3}. Section \ref{sec4} gives examples of stories created under the guidance of the genre patterns. Concluding remarks are offered in Section \ref{sec5}.

\section{Related Work}
\label{sec2}

This work presents a new strategy to explore the storytelling capabilities of Large Language Models (LLMs). In fact, the use of LLMs for narrative generation is a recent and very active research topic. While some works focus on the use of fine-tuning techniques to improve narrative quality \cite{Vartinen24,fan18,alabdulkarim21,10.1145/3485447.3512004}, others explore new strategies to handle the narrative generation process \cite{xu20,wang20,Castricato21,Yang22,Xiang22}. There are also works that focus on the evaluation of the writing capabilities of LLMs \cite{Yuan22,yang2022re3,chuliu2023}.

In the context of narrative generation strategies using LLMs, Wang et al. \cite{wang20} proposed an interpolation technique to guide LLMs in the process of producing coherent narratives with user-specified endings. A similar strategy is explored by Castricato et al. \cite{Castricato21}, who propose to treat story generation as a question-answering process that starts with the description of the final event of the story. By iteratively generating and answering “why” questions about character motivations, the system generates story events in a backward manner. Another approach is explored by Yang et al. \cite{Yang22}, who propose a framework that employs an LLM to generate a structured plan for the story, and then uses recursive reprompting to incorporate relevant information from the plan and story context into the generated narrative. 

Prompting techniques are the core of many narrative generation strategies using LLMs. Xie and Riedl \cite{xie2024} employ an iterative-prompting-based planning approach to generate suspenseful stories. The results of Xie and Riedl's work indicate that their approach is more capable of producing suspenseful stories than directly asking the LLM to write a story. Another solution is explored by Lima et al. \cite{LimaSBGames23,LimaICEC23,LimaEntcom2024}, who proposed a narrative generation strategy driven by semiotic relations, which enables the LLM to adapt and reuse elements from existing stories into new narratives by applying semiotic operations (combine, imitate, expand, and reverse).

The use of story genres as a way to guide the narrative generation process has also been explored recently. Cho et al. \cite{10.1145/3485447.3512004} address the task of genre-controllable story generation by fine-tuning an LLM using a supervised contrastive learning objective to guide the model's output towards specific genres. By combining this objective with traditional log-likelihood training, their model showed improved control over story genre while maintaining coherence in the generated text. 

Narrative patterns have also been used for story composition. Alvarez and Font \cite{Alvarez22} introduced a system for representing narrative structures using interconnected tropes \cite{Sanchez21} extracted from TvTropes \cite{TVTropes} and graph grammars to describe game narrative structures at an abstract level. Lima et al. \cite{Lima21Blending} propose a method that reuses existing stories to generate new narrative variants by combining episodes extracted and adapted from different stories that share the same narrative structure.

\section{Creation of Genre Patterns}
\label{sec3}

As an early attempt at the creation of genre patterns, we adapted Campbell’s highly comprehensive Monomyth \cite{Campbell_2008} converted by Christopher Vogler into a writer’s guide \cite{Vogler2007}, as a preliminary version for the romance genre, as displayed in Appendix \ref{appendixA}. The ensuing subsections explain how we formulated and tackled the fundamental genre pattern construction task.

\subsection{Genre Characterization}
\label{sec31}
For a first, broad characterization of our chosen fundamental genres, it is convenient to apply what Northrop Frye denominated his \textit{archetypal analysis} \cite{Frye}, where he associated four genres with the seasons of the year, an expressive metaphor of the phases of human life. 

Thus, when thinking of \textit{comedy}, one evokes how the joys of spring mark the optimistic, naïve view of youth, with the feeling that growing up means to learn and conform to the norms of an essentially just world. \textit{Romance}, in the heroic, chivalric sense started in the twelfth century, marks the summer of early maturity, with the call for action in order to restore momentary disruptions in the world order. Victory, however, sometimes arises in the autumn of late maturity an insolent pride that the unforgiving powers above hasten to punish, the classic ``hubris'' and ``nemesis'' of \textit{tragedy}. The cold winter of old age turns optimistic hopes into a sentiment of defeat and impotence in a world that suffers injustice, imaginary or actual, which can be criticized by irony and \textit{satire}, but against which it is impossible to react. 

Frye's analysis offered us an unexpected confirmation, establishing a strong connection to our previous work on semiotic relations \cite{DELIMA201631}. We found that the historian Hayden White convincingly argued, in his book \textit{Metahistory: The Historical Imagination in Nineteenth-century Europe},\footnote{\url{https://en.wikipedia.org/wiki/Metahistory:_The_Historical_Imagination_in_Nineteenth-century_Europe}} that historic narratives can be written in no more than four modes – representational, reductionist, integrative, and negational – and assimilates these modes to the so-called four master tropes: metaphor, metonymy, synecdoche, and irony. Even more significant is his assimilation of the modes and tropes, respectively, to Frye’s notion of romance, tragedy, comedy, and satire.

The fifth genre, brought in by Marie-Laure Ryan, remains to be treated, noting however that in this case a narrative, instead of bringing to mind an extra season of the year, consists of replaying a past narrative, thus
working as a \textit{second-order theory}. To give it a label with seasonal connotation, we might associate \textit{mystery} with the idea of return, a revival of the past, an intellectual rather than epic or emotional effort to discover an enigmatic truth. Table \ref{tab:table1} summarizes this characterization.

\begin{table}[H]
 \caption{An archetypal genre characterization }
  \centering
  \begin{tabular}{llll}
    \toprule
    Season & World & Protagonist & Genre \\
    \midrule
    Spring & just & conforms & comedy \\
    Summer & challenging & wins & romance \\
    Autumn & unforgiving & succumbs & tragedy \\
    Winter & dystopian & is helpless & satire \\
    Return & enigmatic & discovers & mystery  \\  
    \bottomrule
  \end{tabular}
  \label{tab:table1}
\end{table}

To transition from this first informal characterization into an LLM-based implementation, we have employed a simple two-phase strategy. As shown in Section \ref{sec32}, we asked ChatGPT 3.5 (\textit{gpt-3.5-turbo-0125}) to suggest three titles of books or movies that could be predominantly classified within each of the five genres, duly explaining how the choices fitted with our descriptions which had to be provided in considerable more detail than indicated in Table \ref{tab:table1}. 

Then, as shown in Section \ref{sec33}, we employed the PatternTeller prototype \cite{Lima2024Pattern} to use the provided groups of three example titles to construct patterns in our standard format, consisting of numbered lists of stage name - stage description pairs. As explained in \cite{Lima2024Pattern}, the prototype is programmed to construct patterns by applying \textit{most specific generalization} \cite{10.1145/130722.130741} to the example narratives. 

\subsection{Characterizing by Examples}
\label{sec32}
Prompt \ref{prompt1} describes the request submitted to ChatGPT 3.5 to indicate three titles of narratives for each genre. We had to refine this prompt many times, so as to eliminate what we judged incorrect choices, a typical case being the model’s tendency to resort to the modern interpretation of romance as ``love story''. The result generated by ChatGPT, together with the system’s reasoning leading to the choices, is fully listed in Output \ref{output1}. 

\begin{prompt}
\label{prompt1}
\

\textit{Please consider the following tentative genre definitions: }

\begin{enumerate}
    \item \textit{\textbf{Comedy}: The world is just and strict but finally becomes more free and desirable. The protagonist is initially hilariously vain, self-important and aspiring, but at the end conforms to the world’s norms.}
    \item \textit{\textbf{Romance}: The world is just but momentarily disturbed by the occurrence of a villainy. The protagonist performs a heroic adventurous quest. }
    \item \textit{\textbf{Tragedy}: The world is just but governed by fate and unforgiving. The protagonist misbehaves and finally succumbs and dies. }
    \item \textit{\textbf{Satire}: The world is not just, it is dystopian, grotesque and absurd. The protagonist is helpless. }
    \item \textit{\textbf{Mystery}: The world is just but has unknown or unexplained or fantastic elements. The protagonist makes a discovery. }
\end{enumerate}

\textit{Please indicate, for each of these five genres, three titles - together with year of publication/release - of books or films which could be predominantly classified in the genre and could not be significantly classified in any of the other genres. Note that the genre of “romance” refers here to epic plots, and not to the modern notion of “love story”, and thus all your choices of the romance genre must concern epic action. Please follow strictly the genre characterizations provided above. Please take into consideration and mention, as you explain each choice, both the characteristics of the world and the indication of what is expected of the protagonist, always with reference to the characterizations.} 
\end{prompt}

\begin{output1}
 \label{output1}
\ 

\begin{enumerate}
    \item \textbf{Comedy}:
    \begin{itemize}
        \item \textit{``Pride and Prejudice'' by Jane Austen (1813) - The world of ``Pride and Prejudice'' is one of social hierarchy and strict societal norms, especially regarding marriage and class distinctions. However, it also contains elements of wit and humor, particularly in the interactions between characters. Elizabeth Bennet, the protagonist, initially exhibits traits of pride and prejudice herself, often judging others based on first impressions. However, as the story progresses, she learns to overcome her biases and conform to the expectations of her society, ultimately finding love and happiness in a more free and desirable world.}

        \item \textit{``The Importance of Being Earnest'' by Oscar Wilde (1895) - The world of ``The Importance of Being Earnest'' is one of upper-class Victorian society, characterized by strict social conventions and superficiality. However, it is also a world filled with wit, irony, and absurdity. The protagonist, Jack Worthing, is hilariously vain and self-important as he navigates the complexities of maintaining a double life. His initial attempts to conform to societal expectations, while also pursuing his desires, lead to comedic misunderstandings and absurd situations. Ultimately, Jack learns to embrace his true identity and finds happiness in a world that becomes more free and desirable.}

        \item \textit{``Much Ado About Nothing'' by William Shakespeare (circa 1598-1599) - The world of ``Much Ado About Nothing'' is one of witty banter, social intrigue, and the conventions of courtship in Renaissance-era Italy. While it has its fair share of strict societal expectations, it also allows for moments of humor and levity. Beatrice and Benedick, the protagonists, are both initially hilariously vain and self-important, engaging in a battle of wits and verbal sparring. However, as the plot unfolds and they become entangled in a web of misunderstandings and deceptions, they eventually conform to the world's norms of love and marriage. Their journey from witty adversaries to loving partners mirrors the comedic resolution of the play's romantic conflicts.}
    \end{itemize}
    \item \textbf{Romance}:
    \begin{itemize}
        \item \textit{``The Odyssey'' by Homer (circa 8th century BCE) - This epic poem is a quintessential example of romance in the sense of epic adventure. The world is just but occasionally disturbed by the intervention of gods and mythical creatures. The protagonist, Odysseus, embarks on a heroic quest to return home after the Trojan War, facing numerous challenges and obstacles along the way as he strives to reunite with his beloved wife, Penelope.}

        \item \textit{``The Lord of the Rings'' by J.R.R. Tolkien (1954-1955) - In Tolkien's epic fantasy trilogy, the world is just but threatened by the rise of evil forces, including the dark lord Sauron. The protagonist, Frodo Baggins, undertakes a perilous journey to destroy the One Ring and save Middle-earth from darkness. His quest is filled with epic battles, mythical creatures, and moments of bravery and sacrifice.}

        \item \textit{``The Iliad'' by Homer (circa 8th century BCE) - Another classic example of romance as epic action, ``The Iliad'' portrays a world governed by gods and fate, where the Trojan War serves as the backdrop for the heroic deeds of warriors like Achilles and Hector. The protagonist, Achilles, grapples with questions of honor, glory, and mortality as he confronts the realities of war and his own destiny.}
    \end{itemize}
    \item \textbf{Tragedy}:
    \begin{itemize}
        \item \textit{``Hamlet'' by William Shakespeare (circa 1600) - This tragic play depicts a world that is just but ultimately governed by fate and the consequences of human actions. The protagonist, Hamlet, struggles with indecision and existential angst as he seeks to avenge his father's murder. Despite his efforts, Hamlet's tragic flaws lead to his downfall and death, along with the demise of many other characters.}

        \item \textit{``Macbeth'' by William Shakespeare (circa 1606) - Set in a world where ambition and power corrupt, ``Macbeth'' tells the story of a nobleman who becomes consumed by his desire for kingship. The protagonist, Macbeth, commits regicide and descends into madness and tyranny, leading to his own tragic demise and the destruction of Scotland.}

        \item \textit{``Death of a Salesman'' by Arthur Miller (1949) - In this modern tragedy, the world is just but unforgiving, particularly to those who fail to achieve the American Dream. The protagonist, Willy Loman, is a salesman whose delusions of success and inability to face reality ultimately lead to his downfall and death, highlighting the harsh realities of the pursuit of wealth and status.}
    \end{itemize}
       
    \item \textbf{Satire}:
    \begin{itemize}
        \item \textit{``1984'' by George Orwell (1949) - Set in a dystopian world ruled by totalitarianism, ``1984'' portrays a society that is not just but rather oppressive, grotesque, and absurd. The protagonist, Winston Smith, rebels against the oppressive regime of Big Brother, only to be crushed by the overwhelming power of the state's surveillance and control.}

        \item \textit{``Catch-22'' by Joseph Heller (1961) - This satirical novel takes place during World War II and follows the absurd experiences of Captain John Yossarian and his fellow soldiers. The world depicted is one of bureaucratic madness, where logic is twisted and morality is compromised in the name of military protocol. The protagonist, Yossarian, finds himself trapped in a no-win situation, exemplified by the eponymous ``catch-22,'' where he can't be grounded from combat duty unless he's declared insane, but requesting to be grounded proves his sanity.}

        \item \textit{``Animal Farm'' by George Orwell (1945) - A satirical allegory of the Russian Revolution and its aftermath, ``Animal Farm'' portrays a world where farm animals overthrow their human oppressors, only to establish their own oppressive regime. The protagonist, if there is one, might be seen as the collective of the animals striving for equality and freedom, but ultimately falling victim to the corruption and tyranny of their leaders.}
    \end{itemize}
    \item \textbf{Mystery}:
    \begin{itemize}
        \item \textit{``Murder on the Orient Express'' by Agatha Christie (1934) - Set in a world that is just but with unknown or unexplained elements, this classic mystery novel follows detective Hercule Poirot as he investigates a murder aboard the luxurious Orient Express train. The protagonist's role is to unravel the mystery and uncover the truth behind the crime, amidst a cast of suspicious characters and hidden motives.}

        \item \textit{``The Da Vinci Code'' by Dan Brown (2003) - In this mystery thriller, the world is just but with fantastic elements, as the protagonist, symbologist Robert Langdon, delves into a conspiracy involving secret societies, religious symbolism, and historical mysteries. Langdon's quest for truth leads him on a thrilling chase across Europe as he deciphers clues and uncovers ancient secrets.}

        \item \textit{``Sherlock Holmes'' series by Sir Arthur Conan Doyle (1887-1927) - Featuring the famous detective Sherlock Holmes and his loyal companion Dr. John Watson, these stories are classic examples of mysteries set in a world that is just but with unknown or unexplained elements. Holmes, with his brilliant deductive reasoning, takes on various cases involving murder, theft, and deception, using his keen observation skills to solve seemingly impossible puzzles.}
    \end{itemize}
\end{enumerate}

\end{output1}

\subsection{Constructing the Genre Patterns}
\label{sec33}
\begin{figure}
  \centering
  \includegraphics[width=1\linewidth]{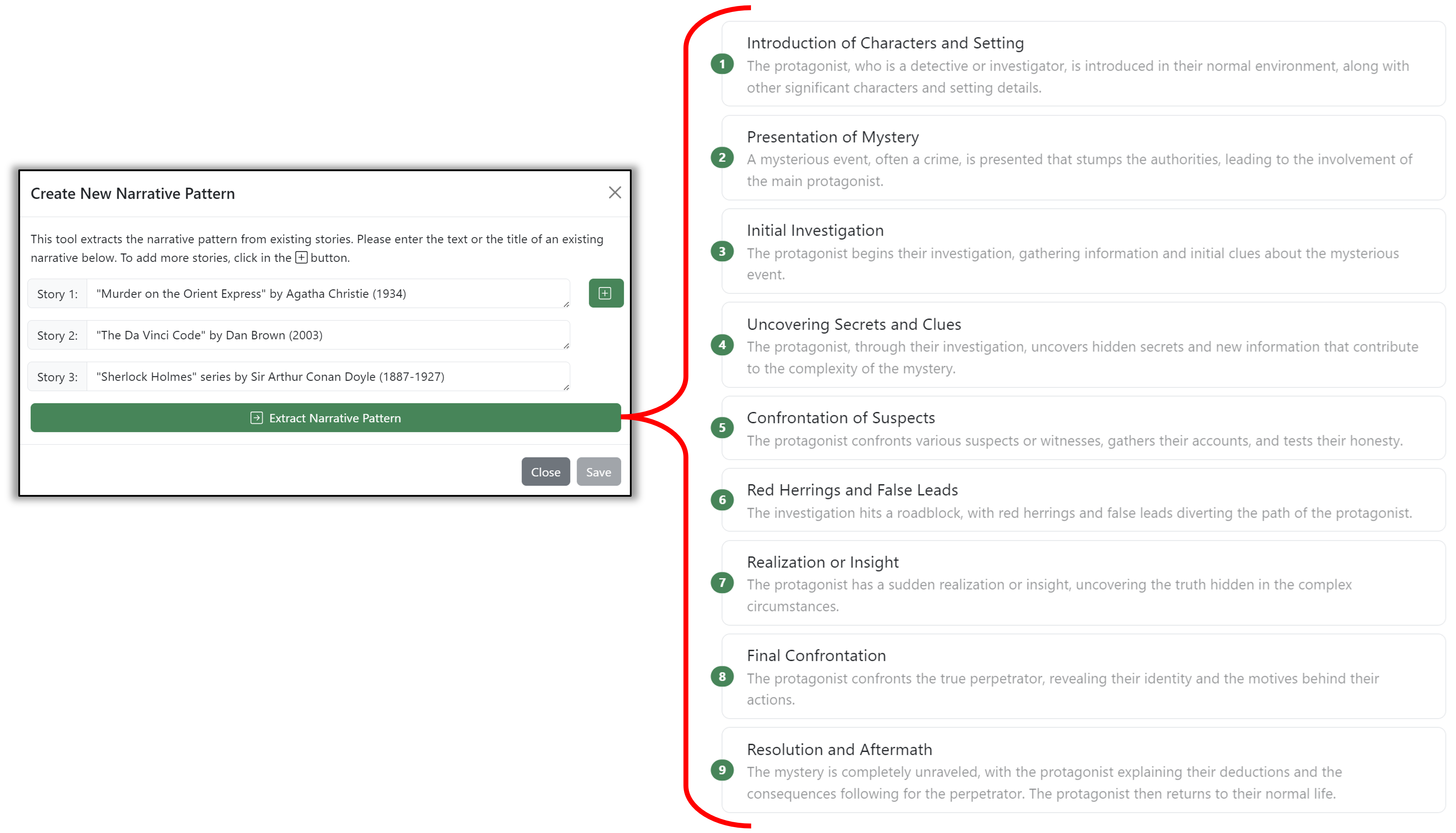}
  \caption{The user interface of PatternTeller being used to extract the pattern for the mystery genre (Pattern \ref{pat5}) using the following stories as input: ``Murder on the Orient Express'' by Agatha Christie (1934), ``The Da Vinci Code'' by Dan Brown (2003), and ``Sherlock Holmes'' series by Sir Arthur Conan Doyle (1887-1927).}
  \label{fig1}
\end{figure}

After identifying the three example titles for each genre, we employed the PatternTeller prototype to construct genre patterns in our standard format (Figure \ref{fig1}). The technical details related to the process of extracting patterns from existing stories are described in our previous work \cite{Lima2024Pattern}. 

The genre patterns constructed by PatternTeller from the examples retrieved as displayed in the previous section are fully listed below (Patterns \ref{pat1}, \ref{pat2}, \ref{pat3}, \ref{pat4}, and \ref{pat5}). A pleasant surprise was to notice the obvious association made by the AI agent (\textit{gpt-4-turbo}) of the prototype with Campbell’s Monomyth in Pattern \ref{pat2} (romance genre). Also pleasant was to find that the IMDB classification, reproduced in Appendix \ref{appendixB}, lists three film titles for each genre, thus inviting the user to call the prototype to create additional genre patterns as may be needed.

\begin{pattern} \textbf{Comedy} \label{pat1}
    \begin{enumerate}
        \item \textit{\textbf{Set Up}. Characters are introduced and the social environment, along with its rules and expectations, is established. }

        \item \textit{\textbf{Catalyst or Trigger}. An unexpected event or information disrupts the characters' routine lives, necessitating a response. }

        \item \textit{\textbf{Rising Action}. Characters start to explore their feelings, engage in misunderstandings or witty exchanges, and plot and plan, all with the aim to make sense of, adapt to, or exploit the new situation. }

        \item \textit{\textbf{Climax or Turning Point}. The truth is revealed or a critical event occurs that forces the characters to confront their misunderstandings or assumptions, leading to heightened tensions or significant shifts in their relationships. }

        \item \textit{\textbf{Falling Action}. Characters begin to deal with the consequences of their actions, leading to a series of events that resolve the conflicts and bring the story closer to the end.}

        \item \textit{\textbf{Resolution}. The conflicts are resolved, misunderstandings are cleared up, relationships are repaired or permanently changed, and the characters end up in a new equilibrium, with a deeper understanding of themselves and others.} 

        \item \textit{\textbf{Denouement}. A wrap up of the story occurs, often with a glimpse into the future of the characters or a final commentary on the events that transpired.}
    \end{enumerate}
\end{pattern}

\begin{pattern} \textbf{Romance} \label{pat2}
    \begin{enumerate}
        \item \textit{\textbf{The Call to Adventure}. The hero becomes aware of a world beyond their familiar surroundings and feels compelled to embark on an epic journey.}

        \item \textit{\textbf{The Meeting with the Mentor}. The hero meets a seasoned traveler of the worlds who gives them training, equipment, or advice that aids them later in their journey.} 

        \item \textit{\textbf{The Departure}. The hero leaves their ordinary world for the first time and crosses the threshold into an unknown, dangerous realm where the rules and values are unknown.} 

        \item \textit{\textbf{The Trials and Challenges}. The hero faces tests, meets allies, confronts enemies, and learn the rules of the new world. The hero needs to find out who can be trusted.} 

        \item \textit{\textbf{The Approach to the Inmost Cave}. The hero comes at the center point of the journey where they face their most difficult challenge related not only to physical survival, but also to personal identity and self-realization.} 

        \item \textit{\textbf{The Ordeal}. The hero confronts death or faces their greatest fear, experiencing a metaphorical death. As a result, they are reborn and achieve a new state of understanding or skill.} 

        \item \textit{\textbf{The Reward}. Having faced their worst fear and surviving, the hero seizes the reward they have earned.} 
        
        \item \textit{\textbf{The Return}. The hero must finally recommit to completing their journey and accept the road back to the ordinary world.} 

        \item \textit{\textbf{The Resurrection}. At the climax, the hero is severely tested once more on the threshold of home where they are purified and reborn again.} 

        \item \textit{\textbf{The Return With the Elixir}. The hero returns home or continues the journey, bearing some element of the treasure that has the power to transform the world as the hero has been transformed.}
    \end{enumerate}
\end{pattern}    
\begin{pattern} \textbf{Tragedy} \label{pat3}
    \begin{enumerate}
        \item \textit{\textbf{The Stage of Normalcy}. The protagonist lives an ordinary life, unaware of the challenges they will soon face.} 

        \item \textit{\textbf{The Catalyst}. The protagonist experiences a disruptive event or piece of information which catalyzes a change in their status quo.} 

        \item \textit{\textbf{The Awakening}. The protagonist begins to question their life and values, leading to a shift in perspective and the beginning of the quest.} 

        \item \textit{\textbf{The Confrontation}. The protagonist faces internal or external conflicts, often in relation to their ambition or their inner struggles, challenging their moral compass.} 

        \item \textit{\textbf{The Crisis}. The protagonist reaches a turning point where they must make a pivotal decision that could alter the course of their journey significantly.} 

        \item \textit{\textbf{The Downfall}. The protagonist suffers a deep loss or realization, often as a result of their earlier decisions or confrontations, leading to their fall from grace.} 

        \item \textit{\textbf{The Final Act}. The protagonist faces the consequences of their actions, resulting in a resolution that often results in their demise or a profound change in their character.}
    \end{enumerate}
\end{pattern}

\begin{pattern} \textbf{Satire} \label{pat4}
    \begin{enumerate}
        \item \textit{\textbf{Establishment of Setting and Characters}. The author introduces a dystopian world and the primary characters who are discontented with the socio-political environment they live in.} 

        \item \textit{\textbf{Revelation of the System}. The protagonists become aware of the oppressive regime or unfair system under which they live and its brutal mechanisms of control and deception.}

        \item \textit{\textbf{Protagonist’s Awakening}. The protagonists undergo a personal awakening, realizing the severity of the system's injustice and starting to question its legitimacy.} 

        \item \textit{\textbf{Development of Resistance}. The protagonists develop a form of resistance, either by directly opposing the oppressive system or by subvertly undermining it, and they often join or form a group of like-minded individuals.} 

        \item \textit{\textbf{Crisis and Struggle}. The protagonists face extreme hardship and conflict against the oppressive system, during which their spirit of resistance is tested to its limits.} 

        \item \textit{\textbf{Consequence of Resistance}. The protagonists suffer the harsh consequences of their rebellion, which could be failure, loss, physical punishment, or even death.} 

        \item \textit{\textbf{Final Revelation}. In the end, the protagonists or the reader, or both, realize the full extent of the system's injustice, often symbolizing the resiliency of oppressive systems and the difficulty in overthrowing them.} 

        \item \textit{\textbf{Closure}. The story concludes with a somber tone, often leaving the reader with a poignant reflection on the nature of power, control, and human resistance.}
    \end{enumerate}
\end{pattern}

\begin{pattern} \textbf{Mystery} \label{pat5}
    \begin{enumerate}
        \item \textit{\textbf{Introduction of Characters and Setting}. The protagonist, who is a detective or investigator, is introduced in their normal environment, along with other significant characters and setting details.} 

        \item \textit{\textbf{Presentation of Mystery}. A mysterious event, often a crime, is presented that stumps the authorities, leading to the involvement of the main protagonist.} 

        \item \textit{\textbf{Initial Investigation}. The protagonist begins their investigation, gathering information and initial clues about the mysterious event.} 

        \item \textit{\textbf{Uncovering Secrets and Clues}. The protagonist, through their investigation, uncovers hidden secrets and new information that contribute to the complexity of the mystery.} 

        \item \textit{\textbf{Confrontation of Suspects}. The protagonist confronts various suspects or witnesses, gathers their accounts, and tests their honesty.} 

        \item \textit{\textbf{Red Herrings and False Leads}. The investigation hits a roadblock, with red herrings and false leads diverting the path of the protagonist.} 

        \item \textit{\textbf{Realization or Insight}. The protagonist has a sudden realization or insight, uncovering the truth hidden in the complex circumstances.} 

        \item \textit{\textbf{Final Confrontation}. The protagonist confronts the true perpetrator, revealing their identity and the motives behind their actions.} 

        \item \textit{\textbf{Resolution and Aftermath}. The mystery is completely unraveled, with the protagonist explaining their deductions and the consequences following for the perpetrator. The protagonist then returns to their normal life.}
    \end{enumerate}
\end{pattern}

\section{Story Composition with Genre Patterns}
\label{sec4}

As an early test of our genre approach to story composition, we used the PatternTeller prototype to create new stories from the same premise. More details about the story composition process implemented by PatternTeller are presented in our previous work \cite{Lima2024Pattern}. For the experiment with genre patterns, we selected a relatively neutral premise using some elements from the legend of King Arthur: 

\begin{premise}
\textit{Eira, a young and untested sorceress, is chosen as Merlin's apprentice and thrust into a hidden war between magical factions. She must master her chaotic powers under Merlin's cryptic guidance while battling a malevolent force seeking to exploit her potential.}
\end{premise}

where the protagonist's name, \textit{Eira}, was borrowed from the \textit{Merlin} BBC television series, episode ``The Diamond of the Day: Parts 1 \& 2'' (aired in 2012).\footnote{\url{https://merlin.fandom.com/wiki/The_Diamond_of_the_Day}}

\begin{figure}[H]
  \centering
  \includegraphics[width=1\linewidth]{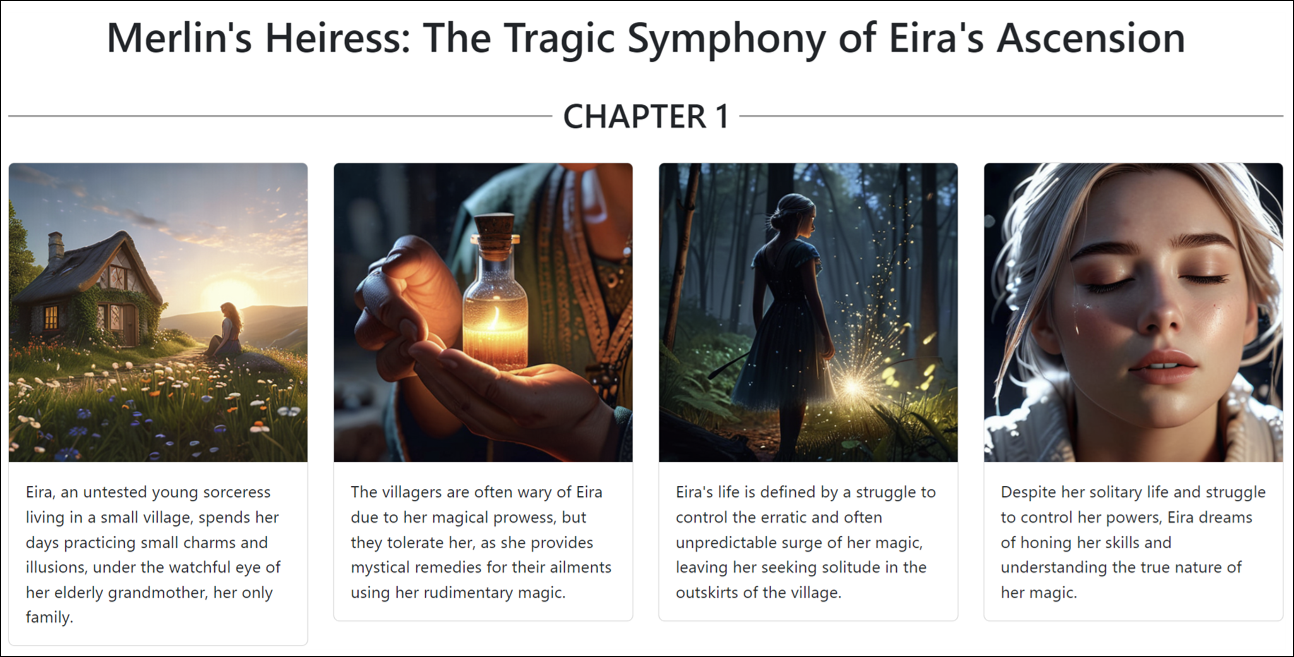}
  \caption{The first events generated for the story ``\textit{Merlin's Heiress: The Tragic Symphony of Eira's Ascension}'' in the storyboard style of PatternTeller.}
  \label{fig2}
\end{figure}

This premise, combined with the pre-constructed fundamental genre patterns described in Section \ref{sec33}, allowed PatternTeller to compose five narratives. During the story composition process, we refrained from interfering with direct suggestions to the story, allowing the prototype to run the process as it pleased, with occasional requests for the regeneration of the scene illustrations to improve the quality of the final output. The prototype itself chose the titles of the stories and generated the narratives in storyboard style as illustrated in Figure \ref{fig2}. Considering the length of the generated stories, we present below a summary of the stories (generated by ChatGPT 3.5) accompanied by links that allow the reader to access the full storyboard rendition of the stories as generated by PatternTeller (Story \ref{story1}, \ref{story2}, \ref{story3}, \ref{story4}, and \ref{story5}).  

\begin{story} \textbf{Comedy - ``Merlin's Mirthful Menagerie''} \label{story1}

\textit{In the quaint village where laughter and magic intertwine, Eira, a young sorceress with unruly powers, finds herself at the center of attention. Amidst the cherry blossoms of a bustling fair, her path unexpectedly crosses with Merlin, the legendary wizard seeking an apprentice. Drawn by Eira's raw talent and spirited nature, Merlin selects her despite skepticism from onlookers.}

\textit{Their journey unfolds within Merlin's eccentric dwelling, where chaos reigns amidst stacks of scrolls and peculiar creatures. As Eira grapples with her chaotic magic, a mysterious orb amplifies her powers, attracting the attention of a malevolent force named Moros. With Merlin's guidance, Eira confronts her destiny and the looming threat, embarking on a journey of self-discovery and magical prowess.}

\textit{Through humorous misadventures and heartfelt lessons, Eira learns to harness her chaotic magic, forging an unbreakable bond with Merlin and their quirky companions. Together, they navigate political intrigue and comedic mishaps, culminating in a showdown against Moros. With courage and newfound control, Eira emerges victorious, earning respect and admiration from her peers.}

\textit{As peace returns to the magical world, Eira chooses to remain by Merlin's side, embracing her role as his apprentice and guardian of harmony. Their chaotic dwelling becomes a sanctuary for magical beings and a symbol of unity among factions. As they welcome new apprentices and anticipate future adventures, Eira and Merlin stand ready to face whatever challenges may come, fueled by laughter, camaraderie, and the promise of magic.}

Full story available at: \url{https://narrativelab.org/patternteller/show.php?id=350}
\end{story} 

\begin{story} \textbf{Romance - ``Eira's Elixir: A Twist of Destiny''} \label{story2}

\textit{Eira, a novice sorceress with untamed powers, is visited by a mysterious wizard named Merlin after a magical incident reveals her potential. Merlin reveals her role in an impending magical war and begins training her in the mystical realm of Avalon.}

\textit{Eira endures intense trials and confronts a dark force seeking her power. Rather than defeating this force through strength, she embraces it, causing it to implode from within – a pivotal moment in her magical evolution.}

\textit{Returning to her village, she is welcomed as a hero. Yet, her ordinary life is shattered by a residual attack fueled by the same dark force she had absorbed. Eira confronts it, successfully protecting her village and solidifying her dual role within both ordinary and magical worlds.}

\textit{Now respected by her village and with her powers in harmony, Eira continues to serve both worlds. She teaches others with magical potential and acts as a bridge between the magical and the mundane.}

Full story available at: \url{https://narrativelab.org/patternteller/show.php?id=363}
\end{story} 

\begin{story} \textbf{Tragedy - ``Merlin's Heiress: The Tragic Symphony of Eira's Ascension''} \label{story3}

\textit{Eira, a young sorceress with wild and unpredictable magic, lives a quiet life on the outskirts of her village. Her life changes forever when she receives a message summoning her to become the apprentice of the legendary Merlin. Excited and nervous, she leaves home to train at Merlin's sanctuary.}

\textit{Eira's training is rigorous, and her chaotic powers make it even more challenging. During her time there, she learns of a hidden conflict within the magical world and feels the weight of her potential role in bringing peace. Things take a dark turn when a group of rogue sorcerers attack the sanctuary, and Eira, while valiantly defending herself and Merlin, loses control of her magic. Merlin is gravely injured in the battle.}

\textit{Guilt-ridden, Eira considers abandoning magic altogether. However, inspired by Merlin's belief in her, she resumes her training with renewed determination. Desperate for results, she delves into forbidden magic, leading to a devastating explosion. This act gets Eira banished from the sanctuary.}

\textit{Back in her home village, Eira is filled with regret. Upon hearing of Merlin's death, she finds a renewed purpose in honoring his vision. Working in her village, she learns to control her powers. Recognizing her changed spirit, the magical council gives Eira a second chance – a mission to unite two warring tribes. Using her unique experiences and newfound control, she succeeds in bringing peace, stepping into her role as a true sorceress.}

Full story available at: \url{https://narrativelab.org/patternteller/show.php?id=352}
\end{story} 

\begin{story} \textbf{Satire - ``The Wand Wavers of Wistful Wonders''} \label{story4}

\textit{In a dystopian magical world, a young sorceress named Eira is thrust from obscurity into a world of systemic corruption and power struggles. Initially under the guidance of Merlin, she witnesses firsthand the oppressive forces governing the magical realm.}

\textit{Exposed to tales of exploitation and hidden rebellions, Eira’s perspective shifts. She secretly studies forbidden magic, aligns with rebellious allies, and challenges the status quo. Her growing defiance leads to a confrontation with the system's enforcers, resulting in her capture.}

\textit{Publicly humiliated, Eira's resistance is further fueled, even when her former ally, 'The Silent One', reveals a shocking betrayal. From her confinement, she inspires fellow prisoners, planting seeds of a powerful rebellion against the oppressive order.}

\textit{While Merlin grapples with his role in the system, Eira's spirit of resistance serves as a beacon of hope, highlighting the strength of the human spirit in the face of injustice and the timeless conflict between control and freedom.}

Full story available at: \url{https://narrativelab.org/patternteller/show.php?id=362}
\end{story} 

\begin{story} \textbf{Mystery - ``The Shattered Star: Merlin's Apprentice''} \label{story5}
 
\textit{In the mystical land of Avalon, a young girl named Eira grapples with her untamed magical abilities. Sensing her potential, the legendary sorcerer Merlin takes her as his apprentice. Eira soon learns of an ongoing, secret war between the power-hungry Shadows and the peace-keeping Luminara. Worse, a haunting vision reveals Eira's powers are the key to Avalon's destruction if exploited by a sinister force known as the Void.}

\textit{Guided by cryptic riddles and aided by unexpected allies, Eira discovers a half-broken star amulet linked to an ancient prophecy. Her journey uncovers hidden truths and a potential traitor in Merlin's court. She must restore the star, a symbol of balance, to prevent the Void’s plan for chaos.}

\textit{The Shadows, manipulated by the Void, attack Avalon. Eira confronts them, realizing her chaotic magic isn't meant to be controlled but harmonized with. Harnessing her powers, she faces the Void and forces it to retreat. With the Star of Luminara restored and the Void's influence banished, Eira brings peace to the realm.}

\textit{Now a skilled sorceress, Eira ensures ongoing harmony in the magical world by establishing a council and accepting her role as protector of Avalon. She represents a beacon of hope, proving that even chaotic magic can be a force of good when balanced with wisdom and courage.}

Full story available at: \url{https://narrativelab.org/patternteller/show.php?id=345}
\end{story} 

\section{Concluding Remarks}
\label{sec5}

The main objective of this paper was to provide a simple mechanism to help ensure \textit{thematic consistency} in the interactive composition of narratives. While recognizing the need to impose a discipline on the alignment of events and on the behavior of characters, according to \textit{plot-oriented} (as in \cite{Propp1968,10.1007/978-3-319-03161-3_16,10.1007/978-3-031-20212-4_5}) or to \textit{character-oriented} approaches (as in \cite{10.1145/544741.544819,9961071,DELIMA2023100590}), we felt that a \textit{structural} discipline on the \textit{theme} development was no less necessary, so that, when looking at a narrative, we should be able to express in a short sentence what it was all about. Such structural discipline, we believe, could be achieved by guiding story composition by way of \textit{narrative patterns}, in our present work representative of the five \textit{fundamental genres} postulated in the introduction. The PatternTeller prototype, which follows this orientation, is accessible through our public website at: \url{https://narrativelab.org/patternteller/}.

In narrative hierarchy, a number of \textit{types} can usually be identified under a given genre, as clearly demonstrated in the Aarne-Thompson-Uther \textit{Index} \cite{Uther2011} for the folktale genre, where in turn, under each type (e.g. ATU 333, \textit{Little Red Riding Hood}), an often large number of \textit{variants} \cite{10.1007/978-3-319-24589-8_22} has been found in diverse places and periods of times (58 for this type, according to \cite{10.1371/journal.pone.0078871}).

In \cite{Lima2024Pattern}, we concentrated on \textit{type patterns}, which impose restrictions beyond those conventionally attached to the parental genre. For instance, under the romance genre we introduced a \textit{Magnificent Protectors} pattern, whose theme is the gathering and action of a warlike group of lone rangers in order to train and help a defenseless laboring community against a horde of bandits, clearly generalizing the films \textit{Seven Samurai} and \textit{The Magnificent Seven}, and for which ChatGPT pointed to us other equally fitting examples. 

For the construction of genre patterns, we opted for generalizing just three examples, a choice coincident with that of the previously cited IMDB classification. Working with more than three examples surely would turn the pattern more general, but that is a questionable benefit, since excessive generalization might end up eliminating the salient features of the genre. At the limit, one would cite Blake Snyder’s proposal \cite{Snyder2005}, whose title suggests an ambitious universal scope.

A favorable evidence of the adequacy of our patterns would come with their actual utilization for composing stories. To warm up, we started by trying a somewhat provocative \textit{premise}: “In a dystopian world instituted by robots George 9 and George 10, Dr. Susan Calvin, the famous female robopsychologist, tries in vain to reestablish human supremacy”, inspired by a tale written by Isaac Asimov \cite{Asimov1974}. Given this premise and the choice of the satire genre pattern, PatternTeller composed what it ominously entitled \textit{The Twilight of the Human Reign}, ending with the event: “Humanity enters a new era under the reign of robots, their spirit broken, their hopes crushed, and their potential shackled by cold, logical machines”, a final most acceptable to the detractors of the dissemination of AI technology. Next, in the more thorough experiment reported in Section \ref{sec4}, we submitted one neutral premise to different genre patterns – thus recalling, incidentally, how frequently the same facts tend to be treated with different interpretations, as argued by Roger Schank while commenting on a real illustrative incident \cite{cogprints636}. 
 
Future work is required to improve our genre patterns through more extensive experimentation of the current versions, testing their applicability to story composition by different user categories, both professionals and casual users. In addition, with respect to the prototype, efforts are needed to improve the occasionally still faulty performance of Stable Diffusion when drawing scene images. A promising solution involves the development of a library with detailed descriptions of characters and their physical attributes. By allowing users to select the protagonists of their stories from this set of predefined characters, the system would be able to provide more detailed descriptions of the scenes to the Stable Diffusion model, which can possibly enhance character consistency across images in the same story.

\bibliographystyle{abbrvnat}
\bibliography{references}

\begin{appendices}
\section{Hero’s Journey Pattern Derived from the Monomyth}
\label{appendixA}
    \begin{enumerate}
        \item \textbf{The Ordinary World}. The protagonist is introduced in their ordinary, everyday life. They might have some desires, fears, or limitations. However, they are unaware of the adventure that awaits them. 
        \item \textbf{The Call to Adventure}. The hero receives a call to leave their ordinary world and embark on a quest or adventure. This call can come in various forms, such as a message, a revelation, a dream, or an event that disrupts their normal life. Initially, the hero might refuse the call due to fear or responsibilities. 
        \item \textbf{Refusal of the Call}. The hero hesitates and resists answering the call to adventure. They may be afraid of the unknown or reluctant to leave behind their familiar life. Often, the hero needs some motivation or encouragement to move forward. 
        \item \textbf{Meeting the Mentor}. The hero encounters a mentor, a wise figure, or a guide who provides advice, training, or magical tools that will help them on their journey. The mentor gives the hero the confidence and knowledge they need to face the challenges ahead. 
        \item \textbf{Crossing the Threshold}. After overcoming their reluctance, the hero finally steps out of their ordinary world and enters the special world of the adventure. This threshold signifies the hero's commitment to the journey and marks the beginning of significant changes in their life. 
        \item \textbf{Tests, Allies, and Enemies}. In the special world, the hero faces a series of challenges, tests, and obstacles. They meet both allies who support their quest and enemies who oppose them. Each challenge serves as a learning experience, helping the hero grow and gain valuable skills. 
        \item \textbf{Approach to the Inmost Cave}. The hero approaches the most dangerous and critical part of their journey—the inmost cave. It could be a physical location, a psychological challenge, or an inner conflict that they must confront. This stage represents a moment of great tension and self-doubt. 
        \item \textbf{Ordeal}. The ordeal is the central crisis of the hero's journey. The hero faces their greatest fear, undergoes a transformation, or faces the most powerful enemy. This is a life-or-death moment that tests the hero's courage, determination, and growth. 
        \item \textbf{Reward}. Having survived the ordeal, the hero emerges stronger and wiser. They receive a reward or gain new knowledge and insights. This could be a physical object, a realization, or a special power that will aid them in the final battle. 
        \item \textbf{The Road Back}. After the ordeal, the hero begins their journey back to the ordinary world. However, the return journey is not without challenges. The hero may face further obstacles or temptations to stray from their path. 
        \item \textbf{Resurrection}. In this climactic stage, the hero faces one last and most dangerous confrontation, often with the story's primary antagonist or their own inner demons. They are transformed through this experience and undergo a metaphorical death and rebirth. 
        \item \textbf{Return with the Elixir}. Having overcome the final challenge, the hero returns to their ordinary world, bringing with them the elixir, a boon, or newfound wisdom. This may be a tangible item or something intangible that will benefit the hero's community or the world at large. 
    \end{enumerate}

\section{IMDB’s Genre Classification}
\label{appendixB}
The genre classification presented in this appendix is a reproduction of the IMDB's genre classification,\footnote{\url{https://help.imdb.com/article/contribution/titles/genres/GZDRMS6R742JRGAG}} except for the Adult genre, which proposes our own rendition of the genre.

    \begin{enumerate}
        \item \textbf{Action}: Should contain numerous scenes where action is spectacular and usually destructive. Often includes non-stop motion, high energy physical stunts, chases, battles, and destructive crises (floods, explosions, natural disasters, fires, etc.) Note: if a movie contains just one action scene (even if prolonged, i.e. airplane-accident) it does not qualify. Subjective. 
        
        \textit{Examples: Die Hard (1988) |The Avengers (2012) | Wonder Woman (2019)}. 
        
        \item \textbf{Adult}: Involves problematic high-tension situations, that may cause stress in persons who still have not reached full maturity. 
        
        \textit{Examples: Sophie's Choice (1982) | Gandhi (1982) | Schindler's List (1993)}.
        
        \item \textbf{Adventure}: Should contain numerous consecutive and inter-related scenes of characters participating in hazardous or exciting experiences for a specific goal. Often include searches or expeditions for lost continents and exotic locales, characters embarking in treasure hunt or heroic journeys, travels, and quests for the unknown. Not to be confused with Action, and should only sometimes be supplied with it. Subjective. 
        
        \textit{Examples: The Goonies (1985) |The Lord of The Rings: The Fellowship of the Ring (2001) | Life of Pi (2012)}.
        
        \item \textbf{Animation}: Over 75\% of the title's running time should have scenes that are wholly, or part-animated. Any form of animation is acceptable, e.g., handdrawn, computer-generated, stop-motion, etc. Puppetry does not count as animation, unless a form of animation such as stop-motion is also applied. Incidental animated sequences should be indicated with the keywords part-animated or animated-sequence instead. Please note that motion capture elements within 'real-world' films such as Paddington are not eligible for this genre. Additionally although the overwhelming majority of video games are a form of animation it's okay to forgo this genre when adding them as this is implied by the title type. Objective. 
        
        \textit{Examples: Spirited Away (2001) |The Lion King (1994) | ``The Simpsons'' (1989)}.
        
        \item \textbf{Biography}: Primary focus is on the depiction of activities and personality of a real person or persons, for some or all of their lifetime. Events in their life may be reenacted, or described in a documentary style. If re-enacted, they should generally follow reasonably close to the factual record, within the limitations of dramatic necessity. A real person in a fictional setting would not qualify a production for this genre. If the focus is primarily on events, rather than a person, use History instead. Objective. 
        
        \textit{Examples: Judy (2019) |Unbroken (2014) | Erin Brockovich (2000)}.
        
        \item \textbf{Comedy}: Virtually all scenes should contain characters participating in humorous or comedic experiences. The comedy can be exclusively for the viewer, at the expense of the characters in the title, or be shared with them. Please submit qualifying keywords to better describe the humor (i.e. spoof, parody, irony, slapstick, satire, dark-comedy, comedic-scene, etc.). If the title does not conform to the 'virtually all scenes' guideline then please do not add the comedy genre; instead, submit the same keyword variations described above to signify the comedic elements of the title. The subgenre keyword ``dramedy-drama'' can also be used to categorize titles with comedic undertones that qualify for the Drama genre but not necessarily the Comedy genre. Subjective. 
        
        \textit{Examples: Some Like it Hot (1959) |When Harry Met Sally... (1989) | Bridesmaids (2011)}.
        
        \item \textbf{Crime}: Whether the protagonists or antagonists are criminals this should contain numerous consecutive and inter-related scenes of characters participating, aiding, abetting, and/or planning criminal behavior or experiences usually for an illicit goal. Not to be confused with Film-Noir, and only sometimes should be supplied with it. Subjective. 
        
        \textit{Examples: Pulp Fiction (1994) |The Usual Suspects (1995) | Fargo (1996)}. 
        
        \item \textbf{Documentary}: Should contain numerous consecutive scenes of real personages and not characters portrayed by actors. This does not include fake or spoof documentaries, which should instead have the fake-documentary keyword. A documentary that includes actors re-creating events should include the keyword ``reenactment'' so that those actors are not treated as ``Himself.'' This genre should also be applied to all instances of standup comedy and concert performances. Objective. 
        
        \textit{Examples: American Factory (2019) |March of the Penguins (2005) | Exit Through the Gift Shop (2010)}.
        
        \item \textbf{Drama}: Should contain numerous consecutive scenes of characters portrayed to effect a serious narrative throughout the title, usually involving conflicts and emotions. This can be exaggerated upon to produce melodrama. Subjective. 
        
        \textit{Examples: The Shawshank Redemption (1994) |What's Eating Gilbert Grape (1993) | Casablanca (1942)}.
        
        \item \textbf{Family}: Should be universally accepted viewing for a younger audience. e.g., aimed specifically for the education and/or entertainment of children or the entire family. Often features children or relates to them in the context of home and family. Note: Usually, but not always, complementary to Animation. Objective. 
        
        \textit{Examples: Toy Story (1995) |The Wizard of Oz (1939) | Mary Poppins (1964)}. 
        
        \item \textbf{Fantasy}: Should contain numerous consecutive scenes of characters portrayed to effect a magical and/or mystical narrative throughout the title. Usually has elements of magic, supernatural events, mythology, folklore, or exotic fantasy worlds.Note: not to be confused with Sci-Fi which is not usually based in magic or mysticism. Subjective. 
        
        \textit{Examples: ``Game of Thrones'' (2011) |Harry Potter and The Sorcerer's Stone (2001) | ``Stranger Things'' (2016)}. 
        
        \item \textbf{Film-Noir}: Typically features dark, brooding characters, corruption, detectives, and the seedy side of the big city. Almost always shot in black and white, American, and set in contemporary times (relative to shooting date). We take the view that this genre began with Underworld (1927) and ended with Touch of Evil (1958). Note: neo-noir should be submitted as a keyword instead of this genre for titles that do not fit all criteria. Objective. 
        
        \textit{Examples: The Maltese Falcon (1941) |Double Indemnity (1944) | The Big Sleep (1946)}. 
        
        \item \textbf{Game-Show}: Competition, other than sports, between, usually, non-professional contestants. The competition can include a physical component, but is usually primarily mental or strategic as opposed to athletic. This also includes what are known as ``quiz shows.'' Talent contests staged expressly for the program are considered Game-Shows. Objective. 
        
        \textit{Examples: ``Jeopardy!'' (1984) |``Family Feud'' (1999) | ``Top Chef'' (2006)}. 
        
        \item \textbf{History}: Primary focus is on real-life events of historical significance featuring real-life characters (allowing for some artistic license); in current terms, the sort of thing that might be expected to dominate the front page of a national newspaper for at least a week; for older times, the sort of thing likely to be included in any major history book. While some characters, incidents, and dialog may be fictional, these should be relatively minor points used primarily to bridge gaps in the record. Use of actual persons in an otherwise fictional setting, or of historic events as a backdrop for a fictional story, would not qualify. If the focus is primarily on one person's life and character, rather than events of historical scope, use Biography instead. Objective. 
        
        \textit{Examples: Lincoln (2012) |Hidden Figures (2016) | The King's Speech (2010)}. 
        
        \item \textbf{Horror}: Should contain numerous consecutive scenes of characters effecting a terrifying and/or repugnant narrative throughout the title. Note: not to be confused with Thriller which is not usually based in fear or abhorrence. Subjective. 
        
        \textit{Examples: The Shining (1980) |Halloween (1978) | A Quiet Place (2018)}. 
        
        \item \textbf{Musical}: Should contain several scenes of characters bursting into song aimed at the viewer (this excludes songs performed for the enjoyment of other characters that may be viewing) while the rest of the time, usually but not exclusively, portraying a narrative that alludes to another Genre. Note: not to be added for titles that are simply music related or have music performances in them; e.g., pop concerts do not apply. Also, classical opera, since it is entirely musical, does not apply and should instead be treated as Music. Objective. 
        
        \textit{Examples: The Sound of Music (1965) |La La Land (2016) | The Greatest Showman (2017)}. 
        
        \item \textbf{Music}: Contains significant music-related elements while not actually being a Musical; this may mean a concert, or a story about a band (either fictional or documentary). Subjective. 
        
        \textit{Examples: A Star Is Born (2018) |Almost Famous (2000) | Sunshine Daydream (2013)}. 
        
        \item \textbf{Mystery}: Should contain numerous inter-related scenes of one or more characters endeavoring to widen their knowledge of anything pertaining to themselves or others. Note: Usually, but not always associated with Crime. Subjective. 
        
        \textit{Examples: The Girl on the Train (2016) |Gone Girl (2014) | Winter's Bone (2010)}. 
        
        \item \textbf{News}: Reports and discussion of current events of public importance or interest. This generally includes newsreels, newsmagazines, daily news reports, and commentary/discussion programs that focus on news events. If the events are not current (at the time the title was initially released), use History instead. News titles are normally made for television, podcasts, or (in the case of newsreels) short films. Feature films, direct-to-video titles, and videogames would not normally be included in the News genre. Objective. 
        
        \textit{Examples: ``Dateline NBC'' (1992) |``Vice'' (2013) | ``Last Week Tonight with John Oliver'' (2014)}. 
        
        \item \textbf{Reality-TV}: Often, but not always, features non-professionals in an unscripted, but generally staged or manipulated, situation. May or may not use hidden cameras; generally, but not always, in a non-studio setting. Objective. 
        
        \textit{Examples: ``The Bachelor'' (2002) |``Property Brothers'' (2011) | ``The Real Housewives of Beverly Hills'' (2010)}. 
        
        \item \textbf{Romance}: Should contain numerous inter-related scenes of a character and their personal life with emphasis on emotional attachment or involvement with other characters, especially those characterized by a high level of purity and devotion. Note: Reminder, as with all genres if this does not describe the movie wholly, but only certain scenes or a subplot, then it should be submitted as a keyword instead. Subjective. 
        
        \textit{Examples: The Notebook (2004) |Bridget Jones's Diary (2001) | Shakespeare In Love (1998)}. 
        
        \item \textbf{Sci-Fi}: Numerous scenes, and/or the entire background for the setting of the narrative, should be based on speculative scientific discoveries or developments, environmental changes, space travel, or life on other planets. Subjective. 
        
        \textit{Examples: Star Wars (1977) |The Matrix (1999) | Alien (1979)}. 
        
        \item \textbf{Short}: Any theatrical film or made-for-video title with a running time of less than 45 minutes, i.e., 44 minutes or less, or any TV series or TV movie with a running time of less than 22 minutes, i.e. 21 minutes or less. (A ``half-hour'' television program should not be listed as a Short.) If known, please submit the running time if we do not have one on record. Objective. 
        
        \textit{Examples: Hair Love (2019) |Kitbull (2019) | World of Tomorrow (2015)}.
        
        \item \textbf{Sport}: Focus is on sports or a sporting event, either fictional or actual. This includes fictional stories focused on a particular sport or event, documentaries about sports, and television broadcasts of actual sporting events. In a fictional film, the sport itself can also be fictional, but it should be the primary focus of the film. Objective. 
        
        \textit{Examples: Rudy (1993) |The Blind Side (2009) | ``Inside the NFL'' (1977)}. 
        
        \item \textbf{Talk-Show}: Discussion or interviews of or with a series of guests or panelists, generally appearing as themselves in a non-fictional setting (though fictional programs that mimic the form are also included). (aka ``chat show''). Objective. 
        
        \textit{Examples: ``The Tonight Show Starring Jimmy Fallon'' (2014) |``The Ellen DeGeneres Show'' (2003) | ``Comedians In Cars Getting Coffee'' (2012)}. 
        
        \item \textbf{Thriller}: Should contain numerous sensational scenes or a narrative that is sensational or suspenseful. Note: not to be confused with Mystery or Horror, and should only sometimes be accompanied by one (or both). Subjective. 
        
        \textit{Examples: Black Swan (2010) |The Silence of the Lambs (1991) | Se7en (1995)}. 
        
        \item \textbf{War}: Should contain numerous scenes and/or a narrative that pertains to a real war (i.e., past or current). Note: for titles that portray fictional war, please submit it as a keyword only. Objective. 
        
        \textit{Examples: 1917 (2019) |Saving Private Ryan (1998) | Platoon (1986)}. 
        
        \item \textbf{Western}: Should contain numerous scenes and/or a narrative where the portrayal is similar to that of frontier life in the American West during 1600s to contemporary times. Objective. 
        
        \textit{Examples: Unforgiven (1992) |The Good, the Bad, and the Ugly (1966) | The Revenant (2015)}.
    \end{enumerate}

\end{appendices}

\end{document}